\documentclass[conference]{IEEEtran}
\usepackage{cite}
\usepackage{amsmath,amssymb,amsfonts}
\usepackage{algorithmic}
\usepackage{graphicx}
\usepackage{textcomp}
\usepackage{xcolor}
\usepackage{url}
\usepackage{listings}
\usepackage{xspace}
\usepackage{cuted}
\usepackage{stfloats}
\usepackage{comment}
\usepackage{subcaption}
\usepackage{multirow}
\usepackage{threeparttable}
\usepackage{array}
\usepackage{soul}

\newcommand{\sts}{Seq2Seq\xspace}

\newcommand{\pmn}{\texttt{PMN}\xspace}
\newcommand{\dialogpt}{DialoGPT\xspace}
\newcommand{\pe}{\texttt{PolyEn}\xspace}

\newcommand{\etal}{\textit{et al.}\xspace}

\newcommand{\tfidf}{\texttt{TF-IDF}\xspace}
\newcommand{\mean}{\texttt{mean}\xspace}

\newcommand{\real}{\mathbb{R}\xspace}

\newcommand{\softmax}{\texttt{softmax}\xspace}
\newcommand{\simi}{\texttt{sim}\xspace}
\newcommand{\identity}[1]{\mathbb{I}({#1})\xspace}

\newcommand{\hr}{\texttt{HR}\xspace}
\newcommand{\hrk}[1]{\texttt{HR@{#1}}\xspace}
\newcommand{\mrr}{\texttt{MRR}\xspace}

\newcommand{\fscore}{\texttt{F1}\xspace}

\newcommand*{\Raju}{\color{black}}
\newcommand*{\e}{\color{black}}
\newcommand*{\nc}{\color{black}}
\renewcommand{\hl}[1]{#1}
\DeclareRobustCommand{\delete}[1]{}

\ifCLASSOPTIONcompsoc
  \usepackage[nocompress]{cite}
\else
  \usepackage{cite}
\fi
\ifCLASSINFOpdf
\else
\fi
\begin{document}
%\title{(TBD) Survey on Persona-Based Conversational AI}
\title{Persona-Based Conversational AI: \\
State of the Art and Challenges}

\author{
\IEEEauthorblockN{
Junfeng Liu\IEEEauthorrefmark{1}\IEEEauthorrefmark{3},
Christopher Symons\IEEEauthorrefmark{2}, and
Ranga Raju Vatsavai\IEEEauthorrefmark{2}\IEEEauthorrefmark{3}}
\IEEEauthorblockA{
\IEEEauthorrefmark{1}
\textit{Lirio AI Research, Lirio LLC}, 
Knoxville, TN, USA\\
\IEEEauthorrefmark{2}
\textit{Behavior Reinforcement Learning Lab, Lirio LLC}, 
Knoxville, TN, USA\\
\IEEEauthorrefmark{3}
\textit{Dept. of Computer Science, North Carolina State University}, 
Raleigh, NC, USA \\
\{jliu, csymons, rvatsavai\}@lirio.com
}
}


% compsoc conferences compared to traditional conferences), use this
%Telephone: (800) 555--1212, Fax: (888) 555--1212}


\maketitle

\begin{abstract}
Conversational AI has become an increasingly prominent and practical application of machine learning. However, existing conversational AI techniques still suffer from various limitations. One such limitation is a lack of well-developed methods for incorporating auxiliary information that could help a model understand conversational context better. In this paper, we explore how persona-based information could help improve the quality of response generation in conversations. First, we provide a literature review focusing on the current state-of-the-art methods that utilize persona information. We evaluate two strong baseline methods, the Ranking Profile Memory Network and the Poly-Encoder, on the NeurIPS ConvAI2 benchmark dataset. Our analysis elucidates the importance of incorporating persona information into conversational systems. Additionally, our study highlights several limitations with current state-of-the-art methods and outlines challenges and future research directions for advancing personalized conversational AI technology.
\end{abstract}


% \fi
\IEEEpeerreviewmaketitle

% main contents
\section{Introduction}
{\Raju Though conversational agents, such as chatbots, have been around for a long time, the practical utilization and effectiveness of these models has increased significantly in recent years due to advances in machine learning and natural language processing.}
Chatbots are widely used in many 
applications today, including automated customer services, personal assistants, healthcare, etc. {\Raju If these technologies are expected to consistently perform at or surpass human level services, then conversational agents will need to be able to adapt to the user's current state and behavior in order to provide more personalized responses. This is particularly important for deployments in the healthcare sector, where conversational topics are typically much more personal and sensitive.}

{\Raju There has been significant research around conversational AI in recent years to expand the capabilities of chatbots and address algorithmic and scaling issues.} Recent methods, such as sequence-to-sequence models (\sts)~\cite{sutskever2014sequence} or transformers~\cite{vaswani2017attention, radford2019language, brown2020language} can be used to capture basic characteristics of a conversation,  including grammar, language flow, etc. {\Raju However, they lack the ability to leverage external resources, such as personal information and behavioral cues that could improve and personalize conversations.}
Recent research efforts have attempted to utilize the power of auxiliary data
that supplements the conversational training, such as 
personas of the speakers~\cite{li2016persona, zhang2018personalizing}, 
the environments in which the speakers are interacting~\cite{mostafazadeh2017image}, 
knowledge-base information~\cite{ghazvininejad2018knowledge}, etc. 

In this paper, we explore recent advances in conversational methods for personalized response generation, and identify limitations and research challenges. In addition, we illustrate how persona information can improve conversations with two state-of-the-art conversational models -
the Ranking Profile Memory Network~\cite{zhang2018personalizing} and the Poly-encoder~\cite{humeau2019poly}. {\Raju Our study also highlights several limitations with current state-of-the-art methods and outlines 
challenges and future research directions for personalized conversational AI.}

\section{\hl{Literature Review}}
\label{sec:literature}
%%%%%%%%%%%%%%%%%%%%%%%%%%%%%%%%%%%%%%%
% Conversational AI
%%%%%%%%%%%%%%%%%%%%%%%%%%%%%%%%%%%%%%%
\subsection{Conversational AI}
% traditional methods
Traditional conversational AI methods require well-structured knowledge (such as a 
knowledge graph), excessive API calls for external dependencies, and human expert
knowledge and intervention for evaluation. 
%These requirements have largely limited the scalability and application scenarios of traditional conversational AI methods. 
These requirements have largely limited the scalability and {\e applicability} of traditional conversational AI \hl{\mbox{methods~\cite{gao2018neural}}}. 

% neural methods
Beyond traditional AI methods, neural  {\e (in particular deep learning-based)} approaches have attracted a lot of interest {\e due to their wide success in the fields of computer vision and natural language processing.}
%from recent researchers due to the power of deep learning.
%
Based on the way the responses are generated, neural conversational models
could be categorized into generation-based methods and retrieval-based methods. 

%%%%%%%%%%%%%%%%%%%%%%%%%%%%%%%%%%%%%%%
% generation-based methods
%%%%%%%%%%%%%%%%%%%%%%%%%%%%%%%%%%%%%%%
\subsection{\hl{Generation-Based Methods}}
\textbf{Generation-based} methods produce responses by generating a
sequence of tokens that is novel to the dataset. 
Sequence-to-Sequence (\sts) models~\cite{sutskever2014sequence} and 
Transformer-based models~\cite{brown2020language} 
are two popular families of generative models.
\sts and Transformer were initially used in machine translation (i.e., mapping a 
sequence of tokens in one language into a sequence of tokens in another language), 
and have achieved great success in many other natural language processing
applications. 
Li~\etal~\cite{li2016persona} proposed a \sts-based Speaker-Addressee model 
that incorporates trainable speaker/addressee embeddings in LSTM encoder/decoder
and trained the model with mutual maximum information (MMI) to address speaker 
consistency and response blandness issues.
Zhang~\etal~\cite{zhang2018generating} developed an Adversarial Information Maximization
model that trains a \sts-based generative adversarial networks (GAN) and a discriminator 
to tell the GAN-generated responses and true responses to promote diversity in 
responses.
Zhang~\etal presented \dialogpt~\cite{zhang2019dialogpt} with an architecture based on GPT-2~\cite{radford2019language} that encodes long-term dialogue history by concatenating all utterances as a long text.
Generation-based methods can provide more creative and novel responses if trained well towards certain objectives. 
However, the unstable quality of the generated responses remains a challenging issue for many generation-based models.

%%%%%%%%%%%%%%%%%%%%%%%%%%%%%%%%%%%%%%%
% retrieval-based methods
%%%%%%%%%%%%%%%%%%%%%%%%%%%%%%%%%%%%%%%
\subsection{\hl{Retrieval-Based Methods}}
\textbf{Retrieval-based} methods, which are also referred to as ranking-based 
models, rank a given set of prescribed candidate responses from the dataset
and then choose the top candidate that matches the input. 
Retrieval models typically learn similarities between the input query and candidates through deep neural network encoders, and then score the candidates based on their similarities to the input query. 
The ranking and training process are similar
to many other typical prioritization tasks.
% {\nc Retrieval models typically adopt deep encoders that learn similarity between the input and candidates through a deep neural network and rank the candidates as other typical prioritization tasks.}
%
\textit{Bi-encoder} architectures~\cite{zhang2018personalizing, dinan2018wizard} 
encode the query and candidates separately
into a low dimensional space before calculating the similarity-based ranking scores.
Bi-encoders are computationally efficient at inference due to their ability to pre-calculate and cache the low dimensional embeddings of the candidates. 
\textit{Cross-encoders}~\cite{wolf2019transfertransfo, urbanek2019learning} learn a joint embedding of each query-candidate pair, typically by concatenating them into a long text and encoding the long text through token-level attention mechanism, then generate a ranking score for the pair based on the joint embedding. 
% \textit{Cross-encoders}~\cite{wolf2019transfertransfo, urbanek2019learning} use a full 
% attention mechanism that combines the input and each candidate into a single long text,
% and attends over each token of the long text, then learns ranking scores given the  {\nc attended embedding} of the combined input and candidates. 
%
Cross-encoder, in general, gains better prediction quality due to better attention over the tokens in the input and candidates, but suffers in terms of computational efficiency at inference time, especially when the candidate set is huge  since the joint embedding could not be pre-calculated without knowing the query.
Humeau~\etal~\cite{humeau2019poly} proposed a \pe that takes advantage of both
bi-encoders and cross-encoders. \pe is able to achieve comparable prediction 
accuracy to the former, and better computational efficiency than the latter 
through caching. 
Retrieval models do not have concerns with response quality as the candidate responses are directly drawn from the existing dataset. 
However, they face the challenge of providing
creative and novel responses outside of the given candidate set. 

Several other methods that were originally designed for document retrieval tasks could also be easily adapted as retrieval-based solutions to dialogue systems.
Luan~\etal~\cite{luan2021sparse} hybridized sparse representations (e.g., bag-of-words)
with the learned dense embeddings to capture both keywords information and higher 
level semantics of the sentences. 
For the learned dense embeddings, they also provided theoretical and empirical proof 
that the longer the texts are,  the larger the embedding dimension it needs to encode the 
semantic information, and proposed a multi-vector method that computes the embedding 
of only the first $m$ tokens in the candidates to gain computational efficiency. 
However, such a multi-vector method is based on a strong assumption that the key 
information appears at the beginning of the text. 
This might hold for document retrieval tasks where the first few sentences usually represent the main idea of the paragraph, but is not necessarily true in conversation data which might consist of short phrases with emphasis at the end. Moreover, multi-vector methods do not gain much efficiency in conversation tasks as the conversation utterances are usually short, compared to document retrieval tasks where the documents usually have long paragraphs. 

There are also some applications that combine generation and retrieval methods. For example, 
Roller~\etal~\cite{roller2020recipes} trained a \pe that first selects
an existing candidate response from the dataset, then appends the retrieve response to
a \sts model to generate a more creative response.
Rashkin~\etal~\cite{rashkin2018towards} built a new Empathetic Dialogue dataset and trained  both ranking-based and generative-based (transformer) models to obtain skills to conduct empathetic conversations.

%%%%%%%%%%%%%%%%%%%%%%%%%%%%%%%%%%%%%%%
% persona-based dialogue systems
%%%%%%%%%%%%%%%%%%%%%%%%%%%%%%%%%%%%%%%
\section{\hl{Related Work}}
\label{sec:related-work}

Many applications of conversational AI require personalized responses. 
\hl{Within the context of this paper, we define a persona as any type of profile containing personal information about a conversational partner that offers context allowing for better understanding of a speaker's meaning/intent or facilitating more appropriate phrasing to improve the likelihood that a speaker's dialogue partner will be more receptive to the information being conveyed. This definition of persona is meant to be broadly inclusive of any type of information that can serve this purpose, including textual descriptions of a person, personal demographic information, past dialog with a conversational partner specifically intended to capture personal characteristics about an individual, or even something as abstract as a machine-learned representation capturing salient aspects of an individual's past behaviors. We note the importance of the persona of both parties in the conversation, not just the persona of the chatbot. In other words, inclusion of the addressee's persona is important to understanding the context of the query and to providing personalized responses.}
\hl{The auxiliary information that a persona provides can offer signals that supplement the conversational context beyond the language itself in the utterances, especially when personalization is desired in a conversation.}
For example,  in precision nudging~\cite{copper2020precision_nudging}, 
the main idea is to provide personalized communication to patients to
encourage them to adopt medically recommended behaviors. 
\hl{\mbox{Zhong~\etal~\cite{zhong2020towards}} and \mbox{Song~\etal~\cite{song2019exploiting}}
also pointed out the critical role that a persona plays in conversational AI to ensure
consistent conversational quality and gain user confidence. 
In particular, a persona is effectively used in various applications
including persona-based chit \mbox{chat~\cite{song2019exploiting, zhang2018personalizing, li2016persona}},
and empathetic \mbox{chat~\cite{zhong2020towards, rashkin2018towards}}, etc.}

\subsection{\hl{Speaker Identity in Conversational AI}}
\label{sec:related-work:speaker-identity}
% speaker-consistency
One of the challenges faced by many research projects is the speaker-consistency 
issue when the model generates inconsistent responses for the same input message.
For example, for the input ``Where are you from?'', the same model might 
respond with ``New York'' or ``London'', as both of the two input-response pairs
might be seen in the training data. 
Li~\etal~\cite{li2016persona} incorporated trainable speaker/addressee embeddings
in the \sts for conversation generation to address speaker consistency issues.
Gu~\etal~\cite{gu2020speaker, gu2021deep} trained a speaker-aware model with 
existing BERT architecture by combining the input word embeddings
with other embeddings (e.g., speaker embedding, segment embedding, etc.) in multi-turn  dialogue response selection tasks. The word embeddings are summed with trainable speaker embedding before feeding into a pre-trained BERT model. 
{\nc These methods address the speaker-consistency issue by utilizing implicit speaker
identity information of the speakers, which relies heavily on the speakers' 
appearances in the whole dataset. 
If a speaker appears in multiple conversations, these methods might be able to learn
the speaker's information from other conversations he or she was involved in.
However, these methods are not able to leverage other conversations for the new speakers or for who appeared in only one conversation in the dataset.} Moreover, these methods are still unable to effectively explore persona information
of the speakers to provide personalized responses, partially due to the lack of 
actual persona data in existing benchmark dialogue datasets, such as
Reddit~\cite{mazare2018training}, Twitter~\cite{ritter2010unsupervised} or
Ubuntu\cite{lowe2015ubuntu} datasets.
%

% speaker-persona
\subsection{Persona-Based Conversational AI}
\label{sec:related-work:persona}
Zhang~\etal~\cite{zhang2018personalizing} made available a Persona-Chat dataset that contains 4$\sim$5 persona profile descriptions for each speaker in the
conversation. This dataset was further extended and used in the NeurIPS ConvAI2 challenge~\cite{dinan2020second}. 
Zhang~\etal also proposed two Profile Memory Networks (\pmn) that 
encode the persona profiles into conversation context for 
generation-based and retrieval-based responses generation tasks, respectively. 
The \textit{Generative \pmn} employs a standard \sts model where the dialogue 
history is encoded with an LSTM and the decoder attends over the embedding of the profile entries. 
For \textit{Ranking \pmn} attends the query embedding over the profile embeddings 
and the scores between the input query and candidates could be calculated for
ranking.
Such attention mechanism could be extended to multiple hops with a  key-value memory network 
over the dialogue history to better encode the conversation context in a multi-turn
dialogue. 
The \pmn methods presented innovative ways to represent the speaker persona profiles  and supplement the conversation context with this additional persona information.
In addition to addressing the speaker-consistency and blandness issues, 
the \pmn methods also suggested ways to explicitly encode the speaker profiles with long-term dialogue history. 
However, encoding the sentences by summing the word embeddings
using the weights from \tfidf is not sufficient to capture high level context
information in the sentences. 
Moreover, the \pmn methods rely specifically on text description of
speakers' profiles and crowd-sourced conversation data based on the profiles. 
In many real applications, it might be very hard to obtain the text descriptions of the users. 
Feature-based (e.g., demographic features) or 
event-based (e.g., users' visit histories) profile data are
probably more common and easy to obtain for many
real-world applications.

% goal of this work
{\Raju In this paper, first, we are interested in exploring if and how persona information can improve the quality of conversation responses with existing state-of-the-art methods. Second, we also provide an analysis of the limitations of current research and identified gaps concerning real-world applications. Finally, we also point out promising research directions to leverage persona information in conversational AI.}
\section{Methods}
In order to compare the effects of persona information in the conversation
tasks, we consider comparing state-of-the-art conversational methods
with and without persona information. 
While generation-based methods are able to generate more novel responses,
it is still a challenging task to evaluate the free-form responses from 
the generation-based methods, especially in terms of personalization.
Existing evaluation metrics, such as BLEU (BiLingual Evaluation Understudy)~\cite{papineni2002bleu}, ROUGE (Recall-Oriented Understudy for Gisting Evaluation)~\cite{lin2004rouge} or Perplexity~\cite{hofmann2013probabilistic} scores,
focus on word overlapping, which might be ignorant of the rare but
important key words that provide personalization.
Retrieval methods, despite their less novel responses, provide an easier
situation for evaluation. 
In this paper, we identified two strong state-of-the-art retrieval-based methods: Ranking \pmn and \pe.
We evaluate the models on the ConvAI2 dataset (described in Section~\ref{sec:experiments:dataset}).

The problem is formulated as: given a input $x = (q, P, H)$,
where $q$ is the query, 
$P=\{p_1, \cdots ,p_j\}$ is the set of text-based persona entries and
$H = [h_1, \cdots ,h_k]$ is the dialogue history, 
the goal is to train a model $f(x, c_i) \rightarrow \mathbb{R}$ that
assigns a score $s_i$ to a candidate $c_i$ from a set of candidate 
responses $C$, then select a best response $c^* = \texttt{argmax}_{c_i\in C} f(x, c_i)$.
\hl{For example, a speaker is assigned with ``I like basketball'' as one of his persona entries ($p_j$), and when he was asked ``What do you do at leisure time?'' ($q$), he might respond ``I watch a lot of basketball games.'' ($c^*$).}

%%%%%%%%%%%%%%%%%%%%%%%%%%%%%%%%%%%%%%%
\subsection{Ranking Profile Memory Network}
%%%%%%%%%%%%%%%%%%%%%%%%%%%%%%%%%%%%%%%
The ranking \pmn first encodes the input query as $q$ like 
other existing word/sentence embedding methods. 
In Zhang~\etal~\cite{zhang2018personalizing}, the embedding of the $i$-th
word of the query $q_i$ is looked up from a trainable embedding layer. 
The sentence embedding is then created in one of two ways. The first option is 
by using the \mean vector of the word embeddings. The second option is by taking the weighted average of the word embeddings with the weights calculated by \tfidf. That is, 
\begin{equation}
\label{eqn:pmn:q}
    q =  
    \begin{cases}
        \frac{1}{l} \sum_i q_i, & \text{if using ``\mean''}\\   
        \sum_i \alpha_i q_i,    & \text{if using ``\tfidf''}
\end{cases}
\end{equation}
where $q_i \in \real^{d}$, $d$ is the embedding size, $l$ is the length of the query, and $\alpha_i = 1/(1+\log(1+\texttt{TF}))$ is the weight of word $q_i$ determined by its inverse
term frequency.
Similarly, each persona entry and the candidate responses are encoded
as $p_j$ and $c$, respectively. 

Then a multi-hop framework is applied around $q$ to generate the context
encoding.
With one hop, the context embedding is the query enhanced with the persona
entries, denoted as $q^+$, which is calculated as 
\begin{equation}
\label{eqn:kvmem}
    q^{+} = q + \sum w_j p_j
\end{equation}
where $w_j$ is the normalized similarity between $q$ and $p_j$ by applying 
a \softmax function over the similarities between $q$ and all $p_j$'s.
The similarity between $q$ and $p_j$ is calculated by a carefully chosen similarity 
function $\simi(q, p_j)$, for example, cosine similarity or dot product.
Then the final response $c^*$ can be generated by sampling from the candidate response 
set $C$ with scores calculated as $s_i = \simi(q^+, c_i)$.

The context embedding $q^+$ could also be extended in multiple hops to encode the dialogue
history through a key-value memory network~\cite{miller2016key}, with the
dialogue history as the keys and the replies as the values.
The multi-hop attention layer enhance $q^+$ by attending over
the keys and output $q^{++}$ is generated in a similar way as in
Equation~\eqref{eqn:kvmem}. With multi-hop, the response is sampled with 
scores $s_i =  \simi(q^{++}, c_i)$.
Note that when number of hops equals zero, the ranking \pmn model compares
the similarity of the query $q$ and candidate $c_i$ directly without
using the persona profiles or dialogue history.

The ranking model is trained with the objective function defined in a strong 
benchmark model Starspace~\cite{wu2018starspace}.

%%%%%%%%%%%%%%%%%%%%%%%%%%%%%%%%%%%%%%%
\subsection{Poly-Encoder}
%%%%%%%%%%%%%%%%%%%%%%%%%%%%%%%%%%%%%%%

\begin{figure*}
    \centering
    \includegraphics[width=.7\linewidth]{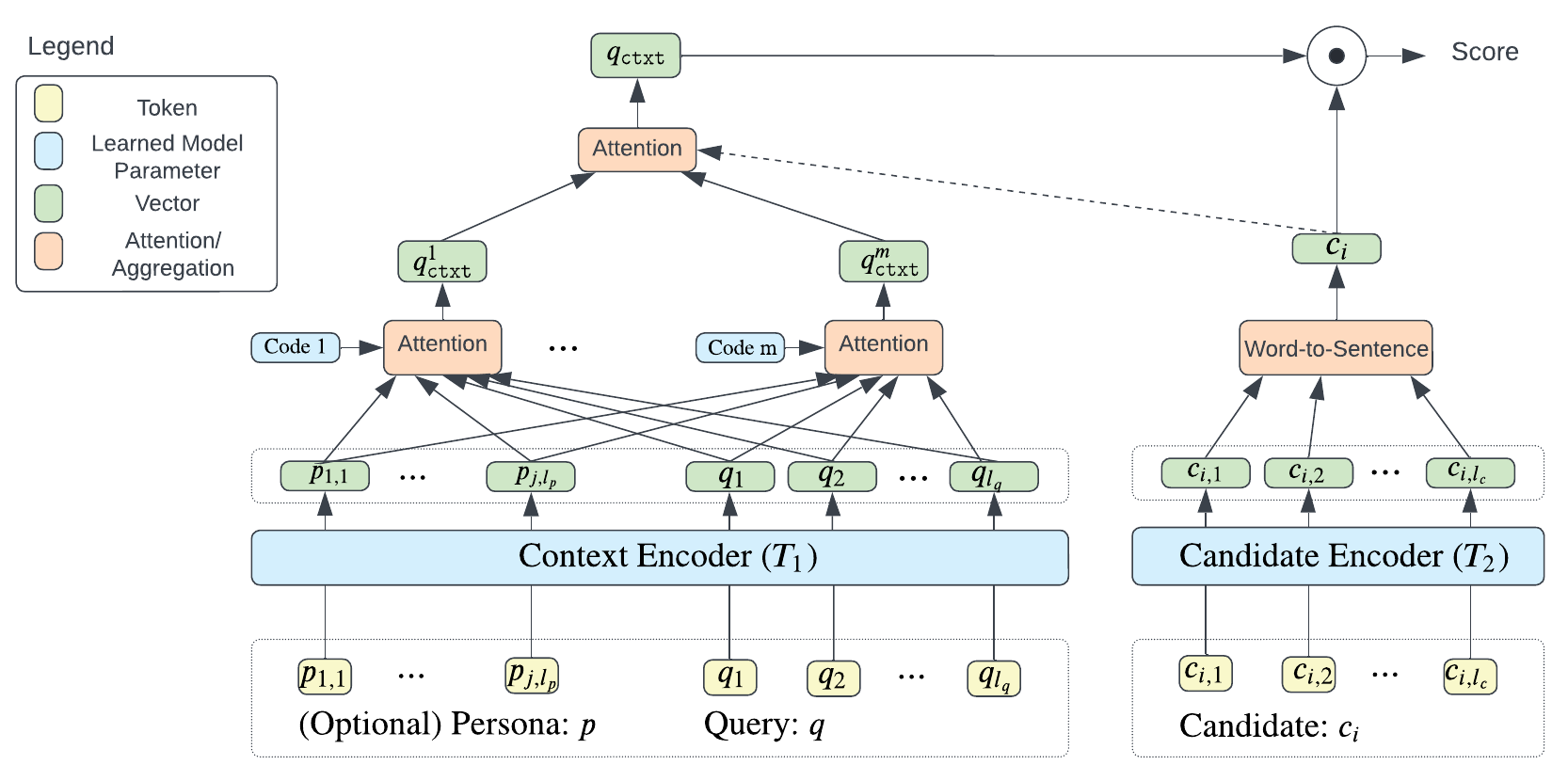}
    \vspace{-5pt}
    \caption{Network Architecture of \pe~\cite{humeau2019poly}}
    \label{fig:network-poly-encoder}
    \vspace{-15pt}
\end{figure*}

The \pe is a ranking-based conversation model that improves 
computational efficiency over Cross-encoders during inference while 
maintains comparable conversation quality to Bi-encoders. 
The network architecture of \pe is demonstrated in Figure~\ref{fig:network-poly-encoder}.
\pe utilizes two separate transformer encoders $T_1$ and 
$T_2$ that encode the query $q$ and the candidates $c_i$'s
separately. One could share the two encoders, i.e., $T_1 = T_2$.
The query and candidate embeddings are represented by the output of the transformer encoders as
\[
    q = T_1(\text{query}), \hspace{10pt} 
    c_i = reduction(T_2(\text{cand})),
\]
where $T(\cdot)$ is the output of the transformer, 
$reduction(\cdot)$ is a reduction function (e.g., mean) along the words,
$q \in \real^{l\times d}$, $c_i \in \real^{d}$, 
$l$ is the lengths of the query,
$d$ is the embedding size.

On the candidate side, the output of $T_2$ is aggregated with a reduction
function to generate $c_i$, a one-dimensional vector representation of the
candidate. As an advantage of Bi-Encoder and \pe, in applications with 
large candidate sets, the candidate embeddings $c_i$ could be pre-calculated
or cached to save significant computational resources at inference time. 

On the context side, the query embedding $q$ is attended over $m$ trainable codes $K = [k_1; \cdots; k_m] \in \real^{m \times d}$ that will generate $m$ global embeddings $q_{ctxt}^1, \cdots, q_{ctxt}^m$ as
\[
    q_{ctxt}^m = \sum_{i=1}^{l}  w_i^{m} q_i
\]
where $q_{ctxt}^m \in \real^{d}$, the attention weights $w_i^{m}$ are derived from the interaction of the $m$-th code and $q$ as
\begin{equation}
\label{eqn:poly:attn-m}
    (w_1^{m},\cdots, w_l^{m}) = \softmax(k_m \cdot q_1, \cdots, k_m \cdot q_l).
\end{equation}
The $m$ global embedding of context $q_{ctxt}^m$ could be viewed as $m$
different points of views to understand the input query, which are controlled
by the $m$ codes. As the model is trained, the $m$ codes will also adjust its
way of viewing the query. 

Then $c_i$ is attended over the $m$ global embeddings which further explores
the relevance between the candidate and the context.
The final context embedding $q_{ctxt}$ is  
\[
    q_{ctxt} = \sum_{i}^{m} w_i q_{ctxt}^i
\]
where $q_{ctxt} \in \real^{d}$ and the attention weights are calculated
similar to Equation~\eqref{eqn:poly:attn-m} as 
$
    (w_1,\cdots, w_m) = 
        \softmax(c_i \cdot q_{ctxt}^1, \cdots, c_i \cdot q_{ctxt}^m).
$
The ranking score of the candidate $c_i$ is calculated as the dot product
with the final context embedding $q_{ctxt}$. The final response is sampled
by scores $s_i =  c_i \cdot q_{ctxt}$.

The original version of \pe itself does not model persona information
explicitly. Given that the ConvAI2 dataset has text-based persona entries,
one could simply concatenate the persona entries with the input query so
that the \pe encodes persona information as a part of the query, i.e., $q \leftarrow [p1; \cdots; p_j; q]$. Similarly,
one could concatenate the dialogue history with the query to encode
the dialogue context, i.e., $q \leftarrow [h1; \cdots; h_k; q]$,  as in Chen~\etal~\cite{chen2020sequential}.

The \pe models is trained to minimize cross-entropy
loss over the logits of the candidates.

\section{Experiments}
\label{sec:experiments}

%%%%%%%%%%%%%%%%%%%%%%%%%%%%%%%%%%%%%%%
\subsection{Dataset}
\label{sec:experiments:dataset}
%%%%%%%%%%%%%%%%%%%%%%%%%%%%%%%%%%%%%%%
We use the dataset from the NeurIPS ConvAI2 competition~\cite{dinan2020second}.
The ConvAI2 dataset is an extended version of the Persona-Chat dataset 
from \pmn~\cite{zhang2018personalizing}. 
The training and testing set of the
Persona-Chat dataset are combined into a larger training set in ConvAI2, 
and there was a new testing set provided for evaluation purposes
during the competition. 
The ConvAI2 dataset was generated by AWS Mechanical Turk tasks. The Turkers (crowdsource workers) were randomly paired up and instructed to
conduct conversations to get to know each other.
The ConvAI2 dataset contains 19,893 dialogues
(17,878 for training, 1,000 for validation and 1,015 for testing). 
Each dialogue has two speakers, and each speaker is assigned with
4$\sim$5 personal profile entries out of a total of 1,155 unique persona
profiles. Each persona profile entry is a short sentence that provides some
information about the speaker, such as ``I like basketball''. 
%
% Zhang~\etal~\cite{zhang2018personalizing} also provided a parallel
% dataset that uses rephrased persona entries to promote generalization, 
% such as ``I love basketball'' or ``I am a sports fan'' instead of specifically % mentioning ``Michael Jordan''.
%

%%%%%%%%%%%%%%%%%%%%%%%%%%%%%%%%%%%%%%%
\subsection{Experimental Setup}
%%%%%%%%%%%%%%%%%%%%%%%%%%%%%%%%%%%%%%%
\label{sec:experiments:setup}
The baseline Ranking \pmn~\footnote{\url{https://github.com/facebookresearch/ParlAI/tree/convai2archive/projects/personachat}} and \pe~\footnote{\url{https://github.com/facebookresearch/ParlAI/tree/main/projects/polyencoder}} models have been implemented with 
ParlAI framework~\footnote{\url{https://parl.ai}}, an open-source 
platform developed by Facebook AI for training and evaluating 
conversational AI models across different tasks. 
We use the ``convai2:self\_original'' task without rephrasing the persona entries.
To make the training process more efficient, we set the training batch size to 64,
and use all the true responses from the training batch as the shared candidate set 
for each query in the batch. 
Humeau~\etal~\cite{humeau2019poly} claimed that a larger training batch size, in 
general, would yield better performance. While we use a smaller batch size compared
to~\cite{humeau2019poly}, we set it the same across all experiments in this paper so that it would not introduce bias to the comparison, and in addition it allowed us to run these experiments on memory-limited GPU nodes.
For the validation set, each input query is 
assigned with a separate set of 20 candidates, among which only 1 is 
the true response. 

\subsubsection{\pe}
We use the pre-trained weights with the Reddit dataset to initialize the model, which contains separate transformer encoders for $q$ and $c_i$'s, 
each with 12 layers, 768 embedding dimensions, and 12 heads in the multi-head 
attention layer. When the persona dataset is used, we simply concatenate the persona 
entries with the input query text and form a long text as the input to the
encoder. Without persona data, the original input query is used as is to serve as the conversation context. 
We try the number of trainable codes $m = $ 5, 16, and 64.
We follow other experimental setup in~\cite{humeau2019poly}.

\subsubsection{Ranking \pmn}
We train the Ranking \pmn model from scratch with the same
architecture as in~\cite{zhang2018personalizing}
with an embedding size of 2000 and cosine similarity between $q$ and $p_i$. 
When persona is used, the $hops$ argument is set to 1, and otherwise, 0. 
The model uses fully trainable word embeddings that are specific to the task. 
As mentioned in Equation~\eqref{eqn:pmn:q}, we try both \mean and
\tfidf as the word-to-sentence aggregation. 

\hl{We denote the \mbox{\pe}/\mbox{\pmn} models trained with persona entries as
$\pe_p$/$\pmn_p$, 
and the \mbox{\pe}/\mbox{\pmn} models trained without persona entries as
$\pe_0$/$\pmn_0$.}

%%%%%%%%%%%%%%%%%%%%%%%%%%%%%%%%%%%%%%%
\subsection{Evaluation Metrics}
%%%%%%%%%%%%%%%%%%%%%%%%%%%%%%%%%%%%%%%
We use evaluation metrics that are commonly used in recommender systems
to evaluate the retrieval-based methods.

The first metric we use is hit rate ($\hr$) at top-$K$ ranking positions,
denoted as $\hrk{k}$. $\hrk{k}$ measures the ratio of the true response 
being ranked in top $K$ by a model in a given batch. 
$\hrk{k}$ is defined as
\[
    \hrk{k} = \frac{1}{|B|} \sum_{(x, c^*) \in B}^{}
              \sum_{i=1}^{k} \identity{c^{r_i} = c^*},
\]
where $B$ is the evaluation batch, $c^{r_i}$ is the candidate 
response being ranked at $i$-th position, and $\identity{\cdot}$
is the identity function which returns $1$ if the expression 
evaluates true otherwise $0$.

The second metric we use is the mean reciprocal rank ($\mrr$), which 
is the reciprocal value of the true response's ranking position in 
the prediction. $\mrr$ of a given batch is defined as 
\[
    \mrr = \frac{1}{|B|} \sum_{(x, c^*) \in B}^{}
              \sum_{i=1}^{|C|} \frac{\identity{c^{r_i} = c^*}}{i}.
\]

We also measure the F-1 score ($\fscore$) of the prediction~\footnote{\url{https://en.wikipedia.org/wiki/F-score}}, which is the 
harmonic mean value of the precision and recall.

Higher $\mrr$, $\fscore$ and $\hrk{k}$ values indicate the model is better
at prioritizing the true responses among the candidates. 
Note that there is only one relevant response among the candidates in the ConvAI2 dataset. Thus, $\hrk{1}$ is equivalent to accuracy, and
the $\hrk{k}$ is equivalent to recall at top-$K$ (\texttt{Recall@k} as reported in
Humeau~\etal~\cite{humeau2019poly}).

%%%%%%%%%%%%%%%%%%%%%%%%%%%%%%%%%%%%%%%
\subsection{Experimental Results}
\label{sec:experiments:results}
%%%%%%%%%%%%%%%%%%%%%%%%%%%%%%%%%%%%%%%

%---------
% Poly-Encoder + PMN Table
%---------
\begin{table*}
\caption{Performance of \pe and \pmn With and Without Persona}
\vspace{-10pt}
\begin{center}
\begin{threeparttable}

\bgroup
\def\arraystretch{1}% 
  \begin{tabular}{
      @{\hspace{10pt}}c@{\hspace{10pt}}|
      c|
      @{\hspace{10pt}}c@{\hspace{10pt}}|
      @{\hspace{10pt}}r@{\hspace{10pt}}
      @{\hspace{10pt}}r@{\hspace{10pt}}
      @{\hspace{10pt}}r@{\hspace{10pt}}
      @{\hspace{10pt}}r@{\hspace{10pt}}
      @{\hspace{10pt}}r@{\hspace{10pt}}
    }

    \hline\hline
    \textbf{Method} & \textbf{Use Persona} & \textbf{Parameter(s)} & \textbf{$\hrk{1}$} & \textbf{$\hrk{5}$} & \textbf{$\hrk{10}$} & \textbf{\fscore} & \textbf{$\mrr$} \\ 
    \hline
    \multirow{6}{*}{\pe} 
        & \multirow{3}{*}{\hl{No ($\pe_0$)}} 
                    &  $m=~5$ &	0.652 &	0.932 & 0.988 & 0.691 & 0.768 \\ 
                &   &  $m=16$ &	\ul{0.674}  &	\ul{0.938} & \ul{0.986} & \ul{0.716} & \ul{0.782} \\
                &   &  $m=64$ &	0.672 &	0.928 & 0.992 & 0.713 & 0.778 \\ 
        \cline{2-8}
        &  \multirow{3}{*}{\hl{Yes ($\pe_p$)}} 
                    &  $m=~5$ &	0.822 &	0.984 & \bf{\ul{1.000}} & 0.840 & 0.890 \\
                &   &  $m=16$ &	\bf{\ul{0.834}} &	\bf{\ul{0.986}} & \bf{\ul{1.000}} & 0.852 & 0.896 \\ 
                &   &  $m=64$ &	\bf{\ul{0.834}} &	0.984 & \bf{\ul{1.000}} & \bf{\ul{0.853}} & \bf{\ul{0.898}} \\ 
    \hline
    \multirow{4}{*}{\mbox{\pmn}} 
        & \multirow{2}{*}{\hl{No  ($\pmn_0$)}}  
                &  \mean    &	\ul{0.295} &	\ul{0.547} & \ul{0.721} & \ul{0.383} & \ul{0.420} \\
            &   &  \tfidf   &	0.275 &	0.529 & 0.725 & 0.353 & 0.405 \\ 
        \cline{2-8}
        & \multirow{2}{*}{\hl{Yes  ($\pmn_p$)}}
                &  \mean    &	0.529 &	\bf{\ul{0.785}} & \bf{\ul{0.889}} & 0.581 & 0.642 \\ 
            &   &  \tfidf   &	\bf{\ul{0.543}} &	0.764 & 0.857 & \bf{\ul{0.592}} & \bf{\ul{0.646}} \\ 
    \hline

  \end{tabular}
  \egroup
          
% \begin{tabular}{ c | c | c | r r r r r}
%     \hline\hline
%     \textbf{Method} & \textbf{Use Persona} & \textbf{Parameter(s)} & \textbf{$\hrk{1}$} & \textbf{$\hrk{5}$} & \textbf{$\hrk{10}$} & \textbf{\fscore} & \textbf{$\mrr$} \\ 
%     \hline
%     \multirow{6}{40pt}{\mbox{\pe}} 
%         & \multirow{3}{15pt}{\hl{No ($\pe_0$)}} 
%                     &  $m=~5$ &	0.652 &	0.932 & 0.988 & 0.691 & 0.768 \\ 
%                 &   &  $m=16$ &	\ul{0.674}  &	\ul{0.938} & \ul{0.986} & \ul{0.716} & \ul{0.782} \\
%                 &   &  $m=64$ &	0.672 &	0.928 & 0.992 & 0.713 & 0.778 \\ 
%         \cline{2-8}
%         &  \multirow{3}{15pt}{\hl{Yes ($\pe_p$)}} 
%                     &  $m=~5$ &	0.822 &	0.984 & \bf{\ul{1.000}} & 0.840 & 0.890 \\
%                 &   &  $m=16$ &	\bf{\ul{0.834}} &	\bf{\ul{0.986}} & \bf{\ul{1.000}} & 0.852 & 0.896 \\ 
%                 &   &  $m=64$ &	\bf{\ul{0.834}} &	0.984 & \bf{\ul{1.000}} & \bf{\ul{0.853}} & \bf{\ul{0.898}} \\ 
%     \hline
%     \multirow{4}{40pt}{\mbox{\pmn}} 
%         & \multirow{2}{15pt}{\hl{No  ($\pmn_0$)} \\ ($hops$=0)}  
%                 &  \mean    &	\ul{0.295} &	\ul{0.547} & \ul{0.721} & \ul{0.383} & \ul{0.420} \\
%             &   &  \tfidf   &	0.275 &	0.529 & 0.725 & 0.353 & 0.405 \\ 
%         \cline{2-8}
%         & \multirow{2}{15pt}{\hl{Yes  ($\pmn_p$)} \\ ($hops$=1)}
%                 &  \mean    &	0.529 &	\bf{\ul{0.785}} & \bf{\ul{0.889}} & 0.581 & 0.642 \\ 
%             &   &  \tfidf   &	\bf{\ul{0.543}} &	0.764 & 0.857 & \bf{\ul{0.592}} & \bf{\ul{0.646}} \\ 
%     \hline
% \end{tabular}
\begin{tablenotes}[flushleft]
            \setlength\labelsep{0pt}
    \footnotesize
    \item Values in \textbf{bold} represent the best performance of the corresponding method irrespective of using persona or not, whereas values with \ul{underlines} highlights best performance of same method with or without persona.
\end{tablenotes}
\vspace{-20pt}
\end{threeparttable}
\end{center}
\label{tbl:performance}
\end{table*}

%Values in \textbf{bold} represent the best performance of the corresponding method. {\nc Values with \ul{underlines} represent the best performance of the corresponding method with the corresponding persona setup. }

\begin{figure*}[!ht]
\centering
\begin{subfigure}{0.31\linewidth}
  \centering
  \includegraphics[width=\textwidth]{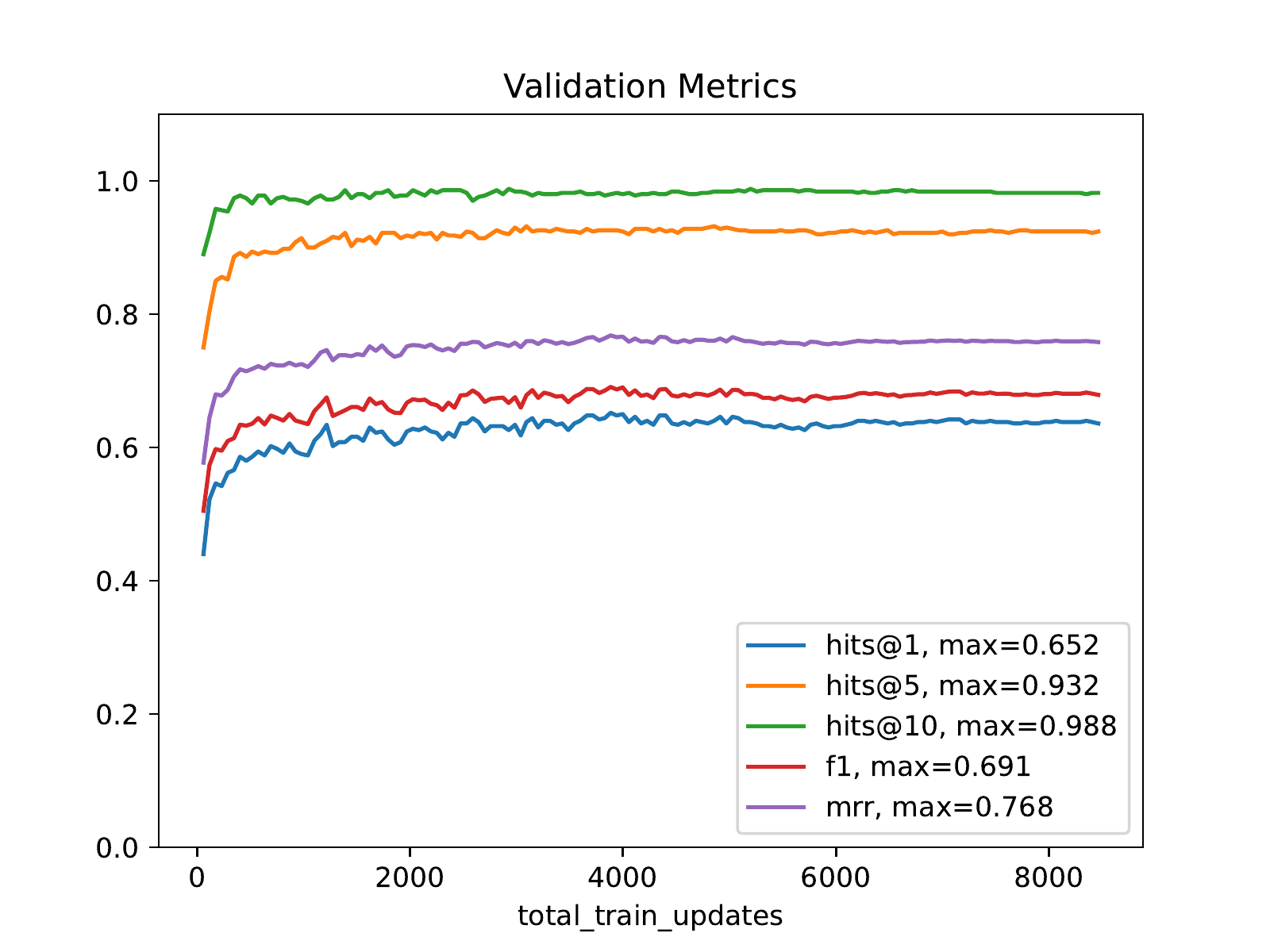}
  \vspace{-15pt}\caption{$\pe_0$, $m = 5$}
  \label{fig:plots-poly-nopersona-m5}
\end{subfigure}
\begin{subfigure}{0.31\linewidth}
  \centering
  \includegraphics[width=\textwidth]{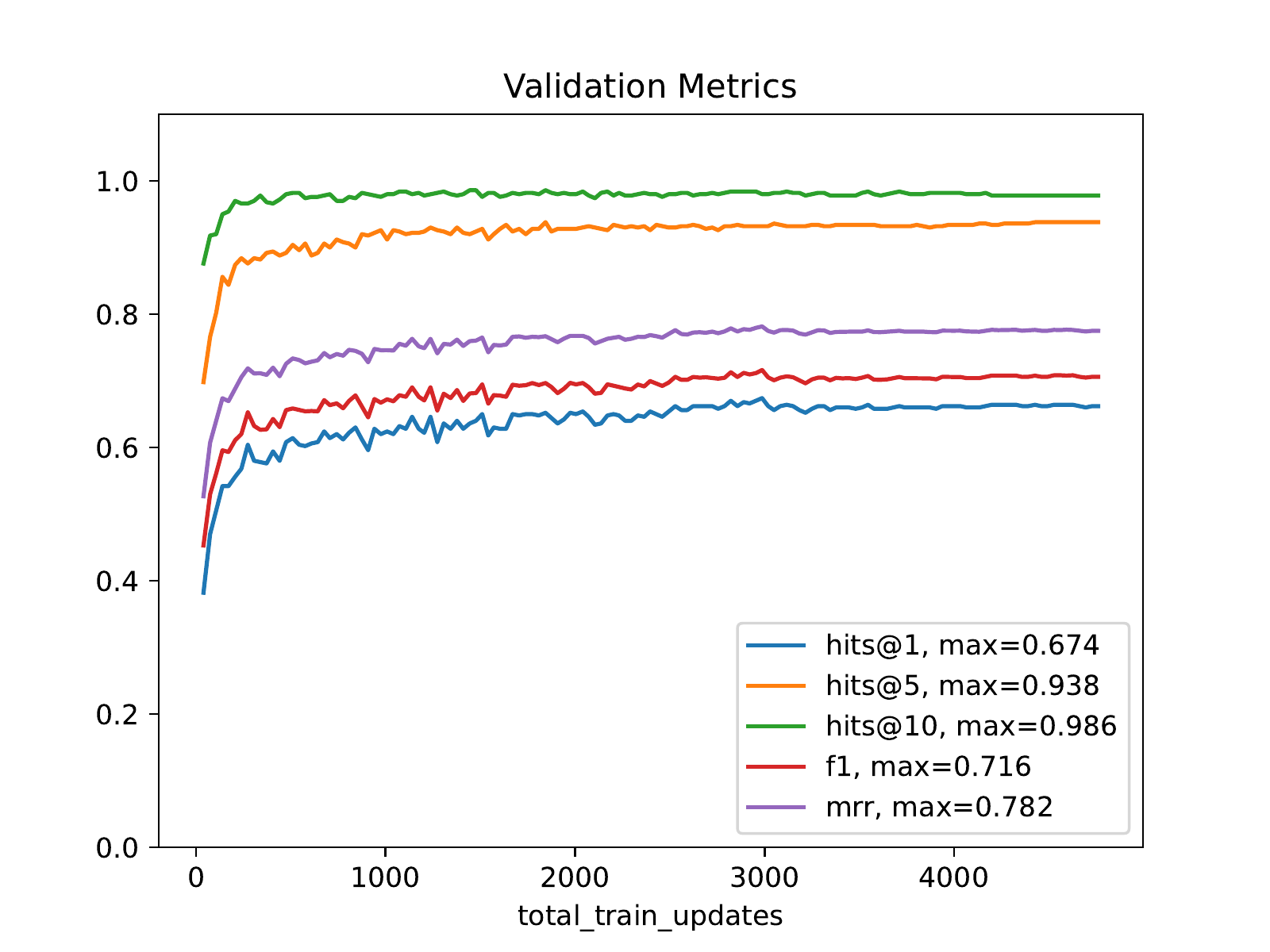}
  \vspace{-15pt}\caption{$\pe_0$, $m = 16$}
  \label{fig:plots-poly-nopersona-m16}
\end{subfigure}
\begin{subfigure}{0.31\linewidth}
  \centering
  \includegraphics[width=\textwidth]{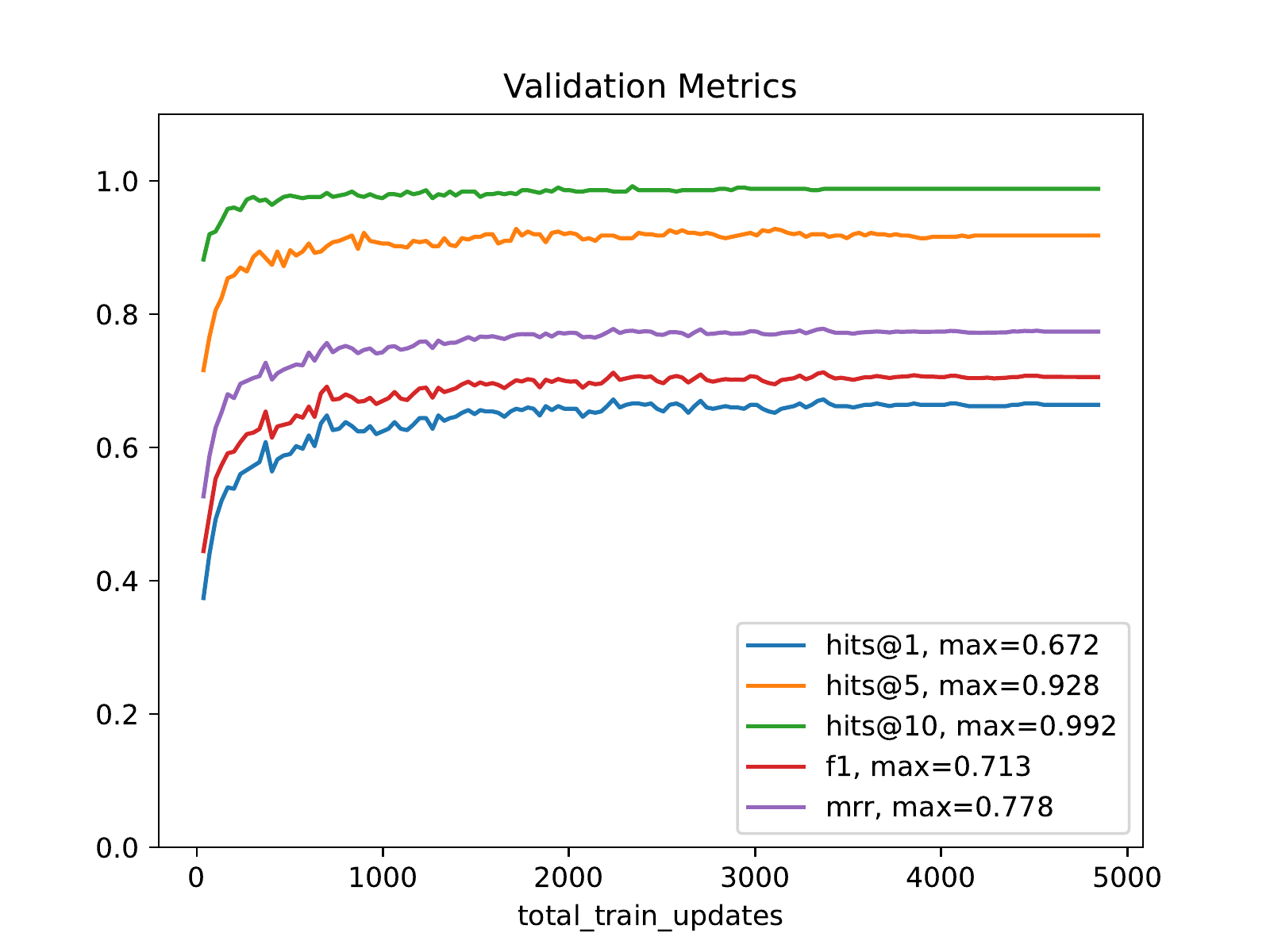}
  \vspace{-15pt}\caption{$\pe_0$, $m = 64$}
  \label{fig:plots-poly-nopersona-m64}
\end{subfigure}
\begin{subfigure}{0.31\linewidth}
  \centering
  \includegraphics[width=\textwidth]{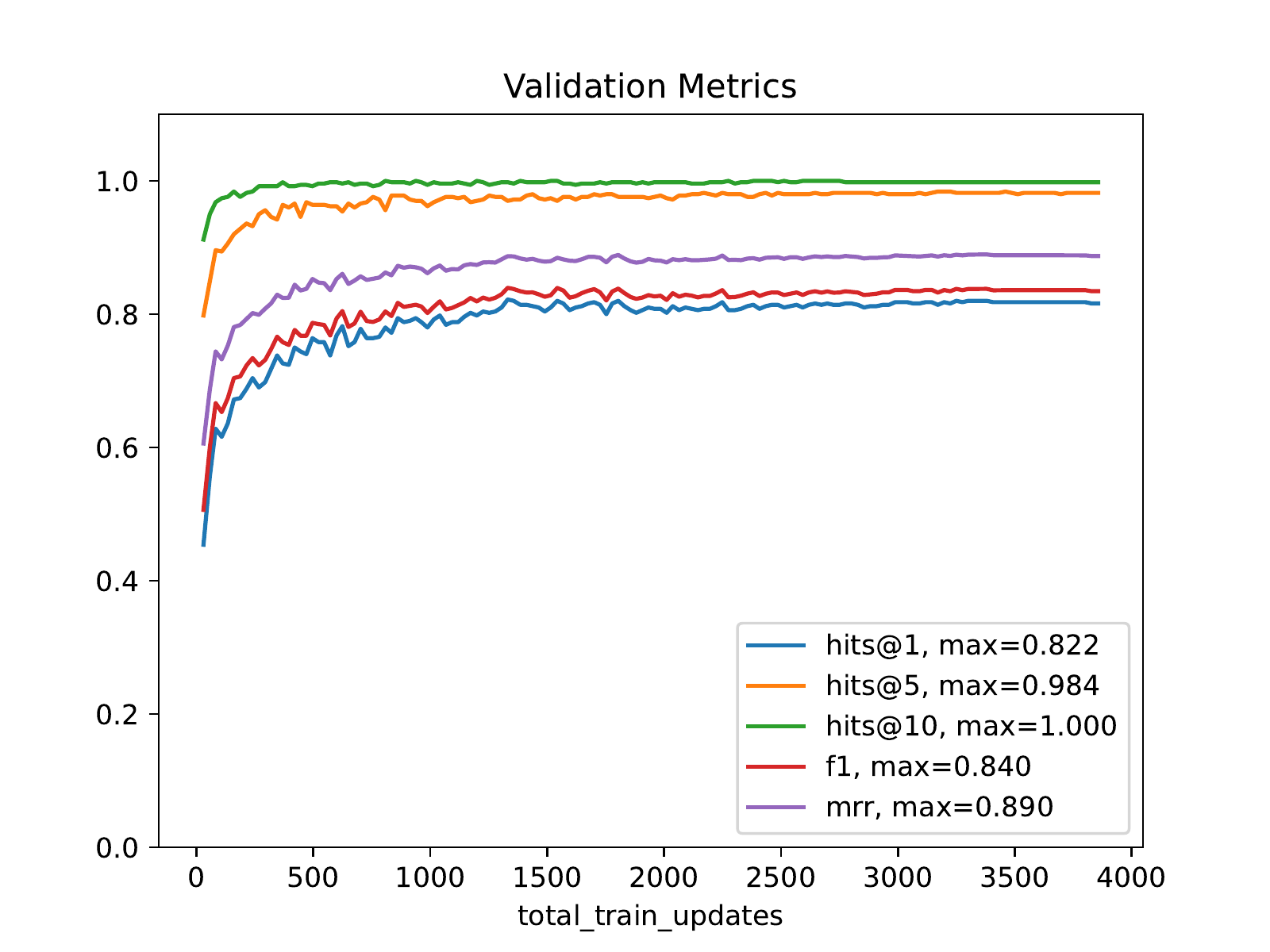}
  \vspace{-15pt}\caption{$\pe_p$, $m = 5$}
  \label{fig:plots-poly-persona-m5}
\end{subfigure}
\begin{subfigure}{0.31\linewidth}
  \centering
  \includegraphics[width=\textwidth]{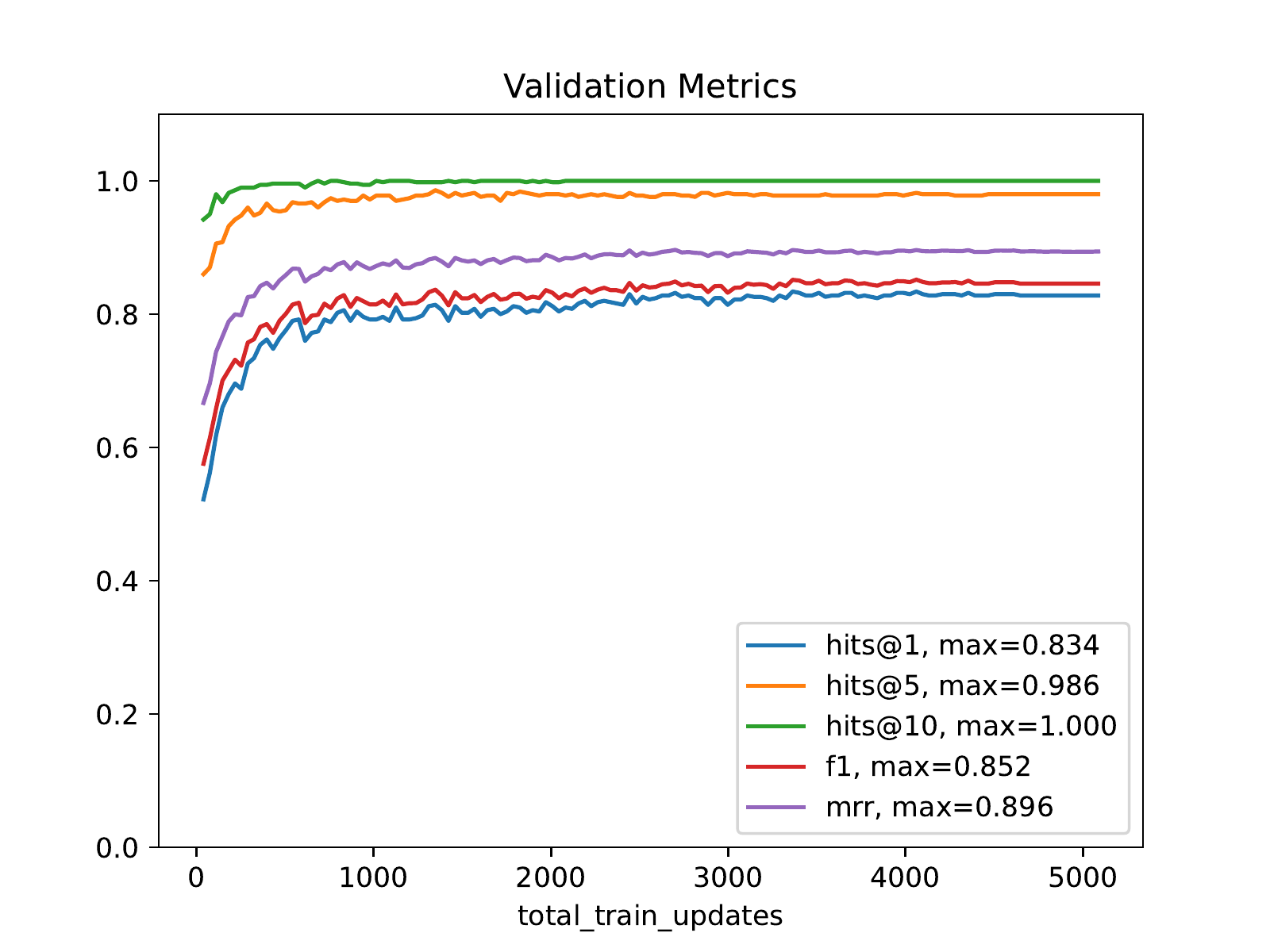}
  \vspace{-15pt}\caption{$\pe_p$, $m = 16$}
  \label{fig:plots-poly-persona-m16}
\end{subfigure}
\begin{subfigure}{0.31\linewidth}
  \centering
  \includegraphics[width=\textwidth]{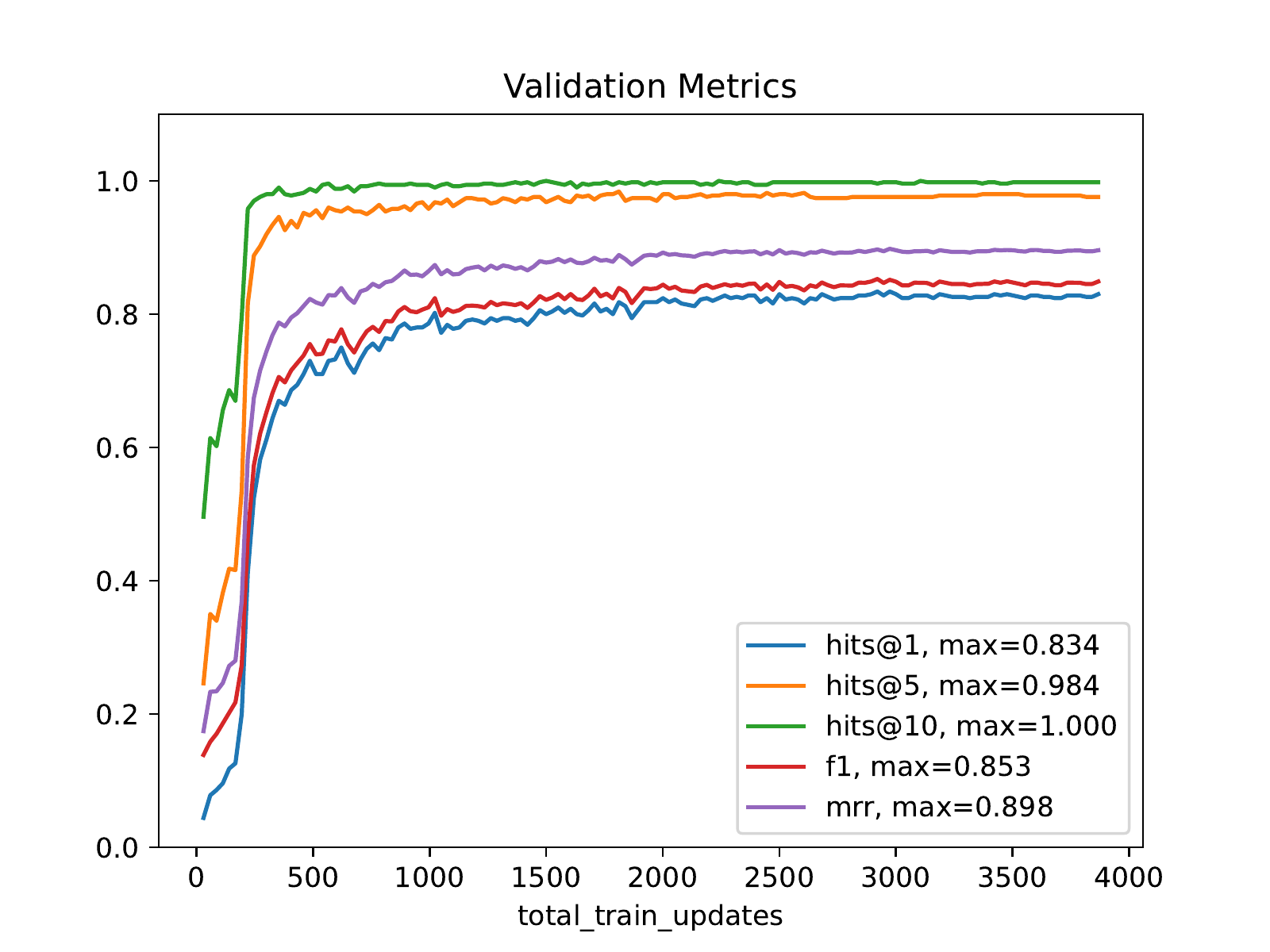}
  \vspace{-15pt}\caption{$\pe_p$, $m = 64$}
  \label{fig:plots-poly-persona-m64}
\end{subfigure}
\caption{Performance on \pe }
\label{fig:plots-poly}
\vspace{-15pt}
\end{figure*}

\begin{figure*}[h!]
\centering
\begin{subfigure}{0.4\linewidth}
  \centering
  \includegraphics[width=0.775\textwidth]{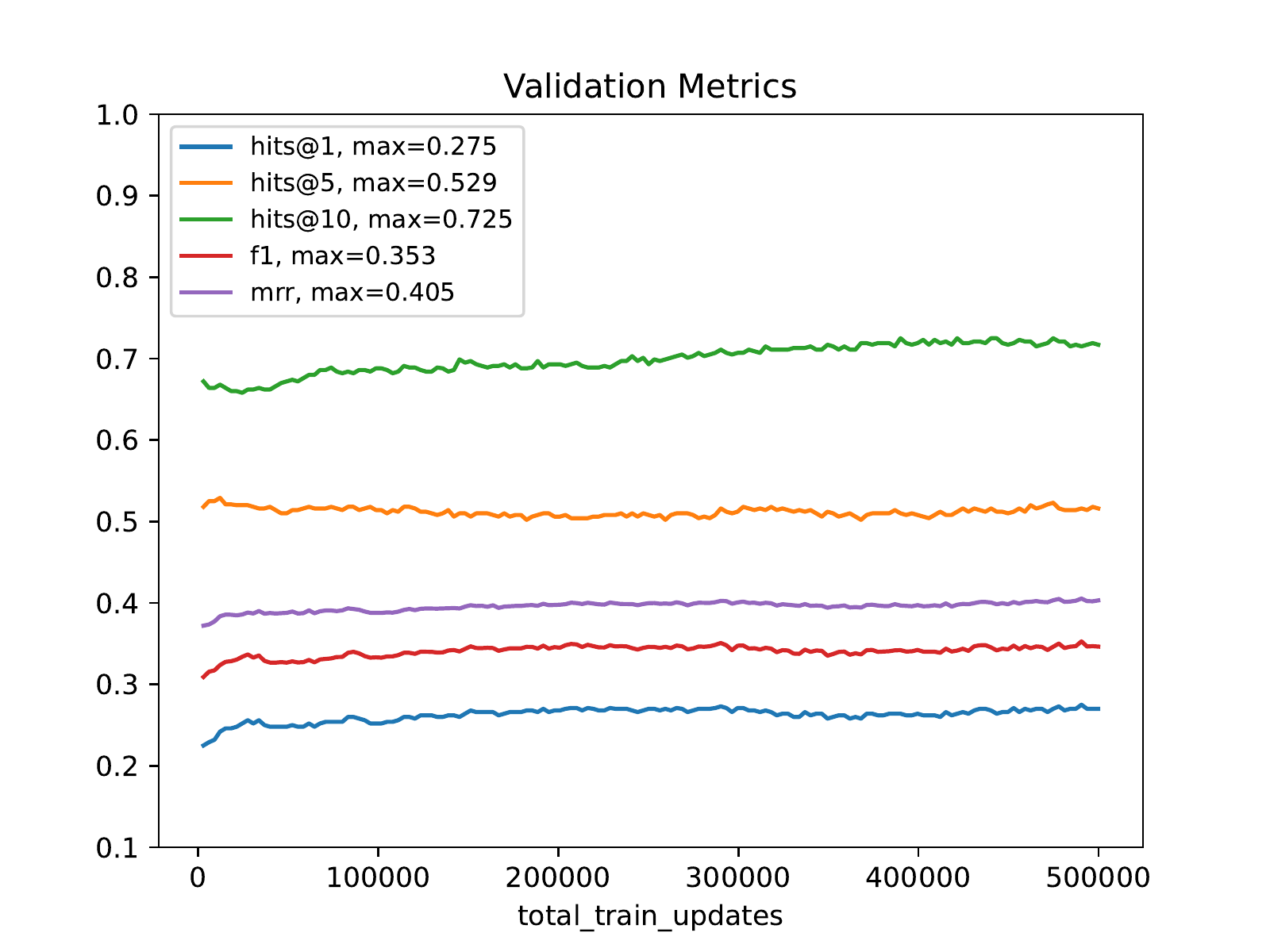}
  \vspace{-5pt}\caption{$\pmn_0$, \tfidf}
  \label{fig:plots-pmn-tfidf}
\end{subfigure}
\begin{subfigure}{0.4\linewidth}
  \centering
  \includegraphics[width=0.775\textwidth]{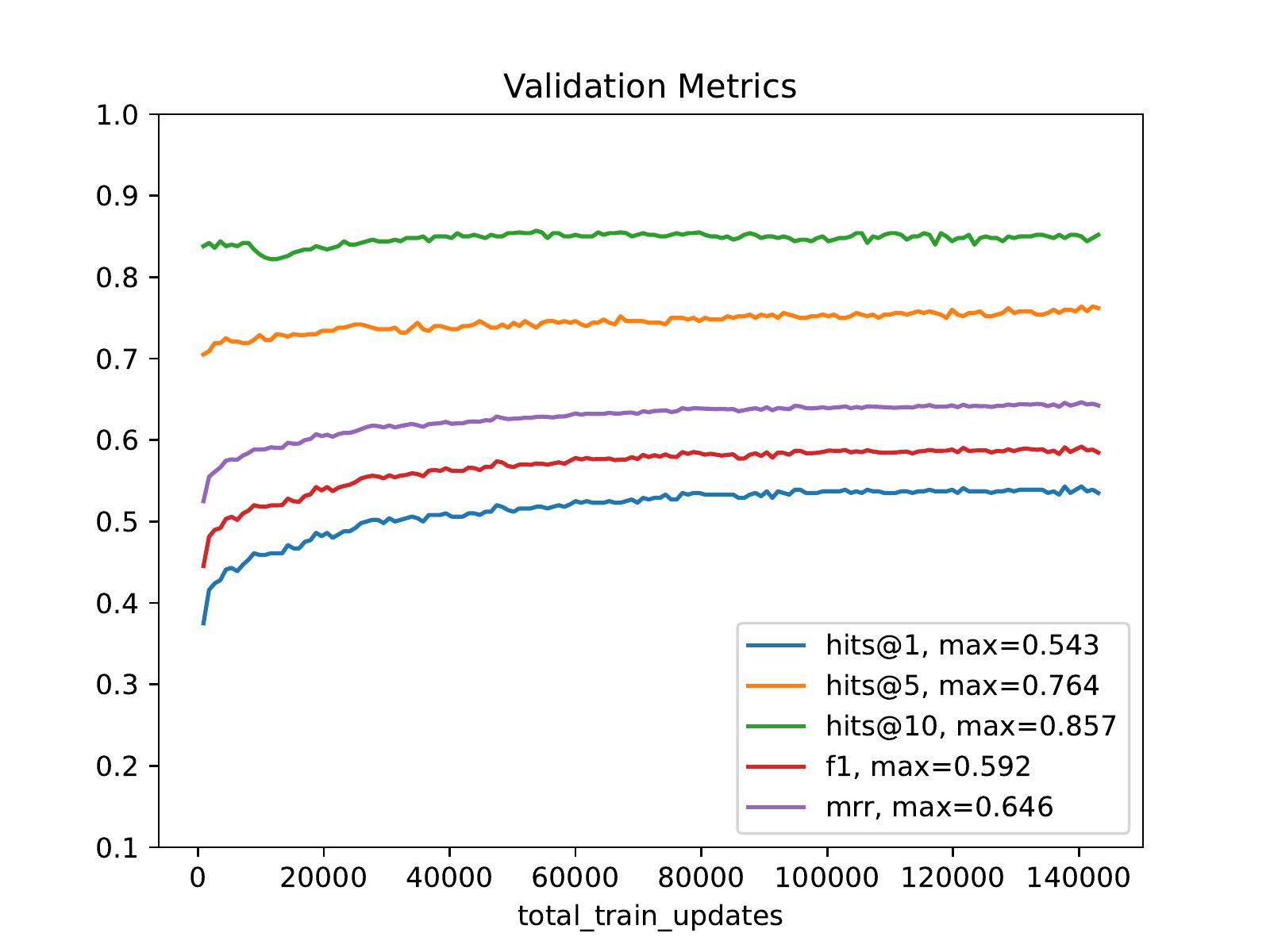}
  \vspace{-5pt}\caption{$\pmn_p$, \tfidf}
  \label{fig:plots-pmn-tfidf-persona}
\end{subfigure}
\begin{subfigure}{0.4\linewidth}
  \centering
  \includegraphics[width=0.775\textwidth]{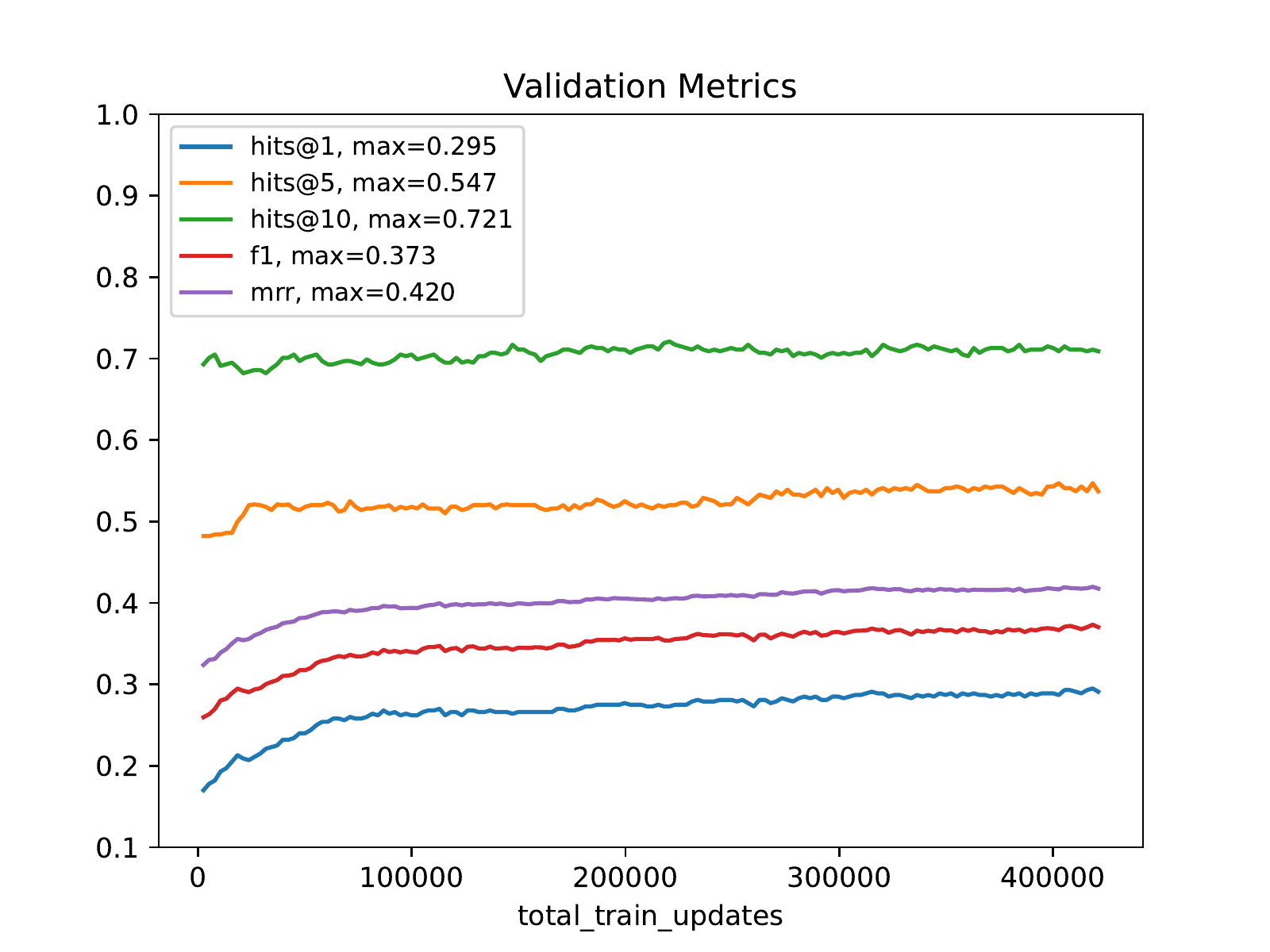}
  \vspace{-5pt}\caption{$\pmn_0$, \mean}
  \label{fig:plots-pmn-mean}
\end{subfigure}
\begin{subfigure}{0.4\linewidth}
  \centering
  \includegraphics[width=0.775\textwidth]{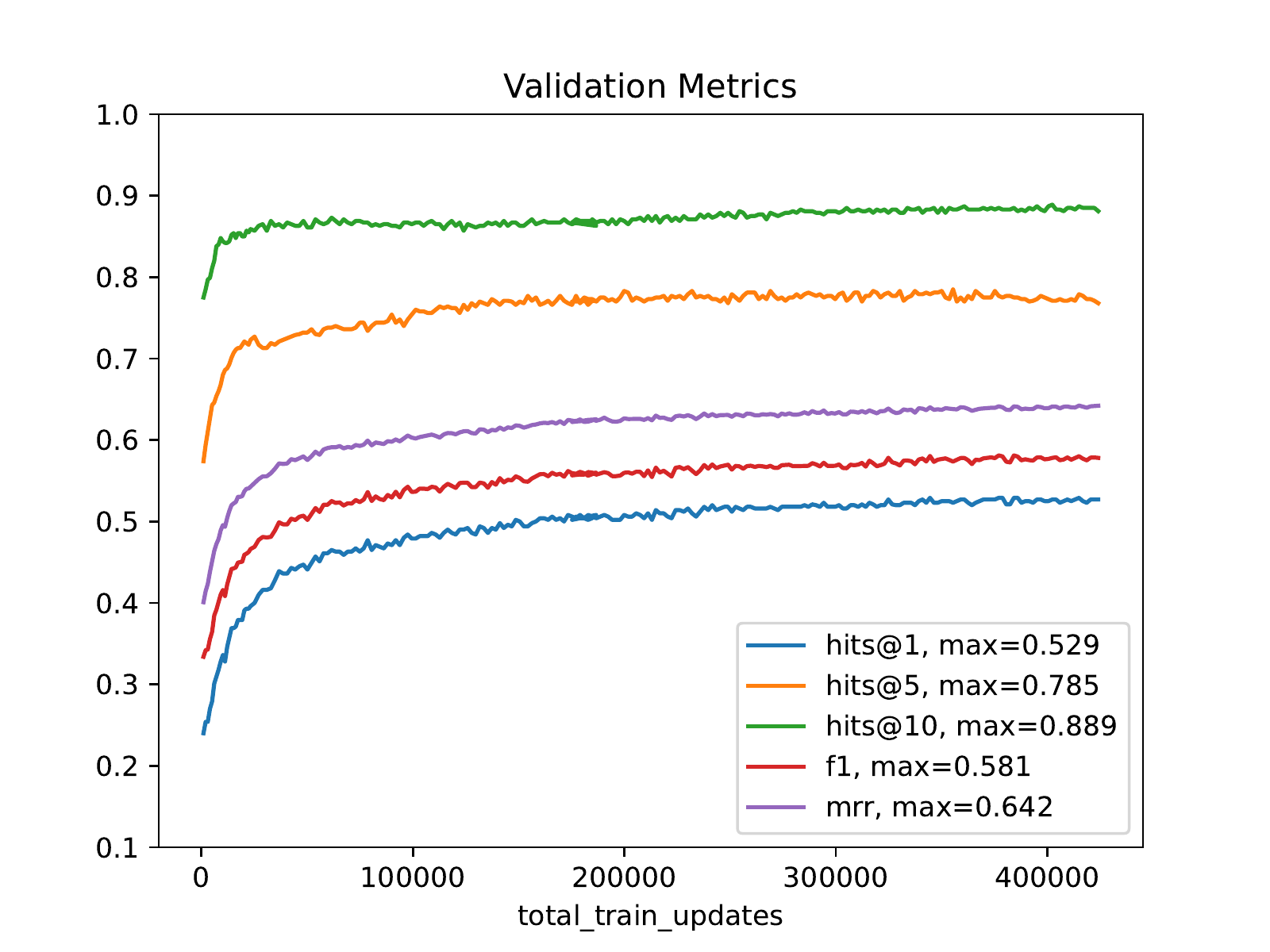}
  \vspace{-5pt}\caption{$\pmn_0$, \mean}
  \label{fig:plots-pmn-mean-persona}
\end{subfigure}
\caption{Performance on Ranking \pmn}
\label{fig:plots-pmn}
\vspace{-15pt}
\end{figure*}

Table~\ref{tbl:performance} compares the performance of the \mbox{\pe and \pmn} methods
with and without persona on the validation set. 
Our experiments showed that both \pe and \pmn  performed significantly better when personas are provided. 
Figures~\ref{fig:plots-poly} and~\ref{fig:plots-pmn} demonstrate the learning curves
of the two methods.

In Table~\ref{tbl:performance}, both \pe and \pmn showed that when personas are used, 
the models are able to prioritize the true responses better than when personas are not 
used, as indicated by the \textbf{bold} values. 
For the $\pe_p$ method, the best \hrk{1}/\fscore/\mrr achieved is 0.834/0.853/0.898 with 
persona and number of code $m=64$. Without persona, the
$\pe_0$ method is only able to achieve 0.674/0.716/0.782.
Similarly, with the $\pmn_p$ method when the persona is incorporated, the model is able to achieve 
0.543/0.592/0.646 with a \mean word vector.
With $\pmn_0$, the metrics drop to 0.295/0.383/0.420.
%
% i don't like reading though
% i don't care for fashion as much as you dislike reading haha
% awesome , i hardly ever read .
%
\hl{In the additional analysis on the validation set, for the best
\mbox{$\pe_p$} ($m$=64) and the best \mbox{$\pe_0$} ($m$=16) models, 
% We evaluated a subset of 5000 validation conversation on both best \mbox{$\pe_p$} and \mbox{$\pe_0$}
% models.
% They respectively achieved \mbox{\hrk{1}} of \textcolor{red}{xx\% (xx) and yy\% (yy)}.
% Among the \textcolor{red}{5000-yy} that \mbox{$\pe_0$} was unable to select the correct responses,
% \textcolor{red}{xx\% (xx)} was correctly predicted by \mbox{$\pe_p$} model. 
%
the \mbox{$\pe_p$} model was able to correct $\sim$64\% of \mbox{$\pe_0$}'s mis-predicted selections,
while \mbox{$\pe_0$} was able to correct only $\sim$38\% of \mbox{$\pe_p$}'s. 
The following is an example. When two speakers are talking about 
hobbies, the query from Speaker 1 is ``I don't like reading though.''
The \mbox{$\pe_p$} model responded ``I don't care for fashion as much as you dislike reading haha,''
since one of persona entries of Speaker 2 (responder) is ``I don't care about fashion'', and it is known to the model.
For the same query, the \mbox{$\pe_0$} model responded ``Awesome, I hardly ever read''
without knowledge of the persona. 
Although the second response is still a legitimate response and captures
a key signal from the query about ``reading'', the first response is more personalized
and better represents the speaker by extracting information from the persona. 
This demonstrates a case where persona entries are able to provide useful signals that supplement conversational context and improve response selection.}

For the \pe methods, we also noticed that the larger $m$ is, the better performance 
the model could achieve. The $\hrk{1}$ performance is 0.822 vs. 0.834 when $m$=5 
and $16$ for $\pe_p$, respectively. This also confirms the conclusions from~\cite{humeau2019poly}. 
The $m$ codes could be interpreted as understanding the sentence from $m$ directions. 
With a larger $m$, it is more likely to find ``proper'' ways to understand the query. 
However, once the value of $m$ is large enough, the performance increase is marginal 
($m$=64 and $m$=360 have very close performance in~\cite{humeau2019poly}).
This is also indicated by our experiments that the performance difference is negligible 
when $m$=16 or $m$=64. This is probably because as $m$ increase, it is already large
enough to provide sufficient information for the model to gain a good understanding of the 
language. Large number of directions ($m$ values above 64) may introduce additional noise,
which is probably why the performance is slightly higher when $m$=16 ($\hrk{1}$=0.674) 
than $m$=64 ($\hrk{1}$=0.672) for $\pe_0$.
%{\nc When excessive directions are inserted (e.g., $m$ value is excessively large), it might inversely provide additional noise to the model, which is probably why the performance is slightly higher when $m=16$ ($\hrk{1}=0.674$) than $m=64$ ($\hrk{1}=0.672$) when the persona is not used.}
%
{\Raju Meanwhile, larger $m$ also means a more complex model with a large number of training parameters requiring more computational resources for training and inference. Therefore, we limited our 
experiments to $m \le 64$.}

{\Raju For the Ranking $\pmn_p$ method, we observed that sentence embedding with the weighted word \tfidf embedding method outperforms naive \mean vector representation ($\hrk{1}$ of 0.543 vs. 0.529).}
This shows that \tfidf embedding is able to provide additional term importance information 
based on the frequencies of the terms compared to the simple \mean encoding. 
It is helpful for the conversational model to capture the relevance between 
the query and the candidates. 
But interestingly, we didn't observe the same trend when the persona is not used. \tfidf did not
outperform the \mean vector encoding for $\pmn_0$ ($\hrk{1}$ of 0.275 vs. 0.295). 
However, both 
\mean and \tfidf have \hrk{1} of less than 0.3. We believe it is because the Ranking \pmn 
model is too simple, and when the information provided to the $\pmn_0$ model is very little (i.e., 
without persona), the model simply could not learn well enough. Such low performance on both 
\tfidf and \mean vector are too low for the comparison to be meaningful. 

The Ranking \pmn method in general does not outperform the \pe method. 
This is largely because the Ranking \pmn method currently only use a very
naive sentence embedding (using \mean word vector or weighted by \tfidf), 
which ignores the order of the words in a sentence. {\Raju As a result the Ranking \pmn method fails 
to capture sufficient signal from higher level context of the language.  One could possibly improve the performance by replacing the naive sentence aggregation with more sophisticated encoders such as an RNN or a transformer. These extensions will be explored in our future work.}

%---------
% Poly-Encoder Plots
%---------
%---------
% PMN Plots
%---------

%\section{Discussions}
\section{Challenges and Opportunities}
%In this section we focus on following key questions

%%%%%%%%%%%%%%%%%%%%%%%%%%%%%%%%%%%%%%%
\subsection{Effectiveness of Persona Information}
%%%%%%%%%%%%%%%%%%%%%%%%%%%%%%%%%%%%%%%
Our experimental results demonstrate that persona information can play an
important role in improving the retrieval performance on both of
the \pe and Ranking \pmn methods. 
The best $\pe_p$ model is able to improve $\hrk{1}$ by 
23.74\% over the best $\pe_0$ model, and improves \fscore
by 19.13\% and \mrr by 14.83\%. 
Similarly, the best Ranking $\pmn_p$ model is able to improve 
$\hrk{1}$, \fscore and \mrr by 84.07\%, 54.57\% and 53.81\%, 
respectively, over $\pmn_0$. 
This illustrates the effectiveness of using the persona files of the speakers
that can help the model understand the conversational context better 
and further improve the performance of a conversational agent. 
It also suggests a promising research direction that 
focuses on better methods to utilize auxiliary information
beyond personas in order to improve natural language understanding
and response generation, such as health information, 
emotion and/or environment of the speaker, socioeconomic data, etc.
{\Raju In particular, the existing methods highlighted in this paper are using an early fusion approach by simply concatenating auxiliary information with input data. This works well if the input data modalities are the same. However, the data modalities that can be used for personalization are diverse and complex, including but not limited to socio-economic attributes, life-style attributes, spatio-temporal attributes. Simple early fusion of these multimodal data streams via concatenation is neither appropriate nor feasible. Further research is required on how to handle multimodal personalization data streams. 
%We are exploring two promising research directions of intermediate fusion and late fusion via a multi-stream network design. %shall we keep last sentence? (Pros: by saying this we are stating that we are already working on this limitation. Cons: are we giving away too much information if I say that here with regard to our future research plans?).
}

%=========================
\subsection{Lack of Realistic Persona Dataset}
%=========================
One main challenge for persona-based conversational AI is 
the lack of a good dataset that reasonably reflects practical
usage. 
% \textcolor{red}{(might need to remove it) {\Raju lets keep it. I moved and added connecting sentences}}
{\Raju From the literature review, we found the following two persona datasets.}
Qian~\etal~\cite{qian2018assigning} constructed a conversation dataset 
from Weibo, a Chinese social media platform, that contains 6 manually extracted 
binary features from the posts. These binary features describe whether
the query/response pair mentioned a certain feature or not (e.g., does
it mention any location), and these were used to train a classification task. 
Such binary features do not contain necessary persona information to
improve personalization in a conversation. 
%

%{\Raju The ConvAI2 dataset which is based on the Persona-Chat dataset~\cite{zhang2018personalizing} was introduced at the NeurIPS competition~\cite{dinan2020second}.} 
Although the ConvAI2 dataset provided a method for associating a conversation with a persona, there is still a large gap between this data and more realistic data that might be available in practical use cases. The ConvAI2 dataset relies on text description-based persona entries, which could be expensive and difficult, if not impossible, to obtain in large scale applications. In some applications, such as healthcare, it could be particularly difficult to collect such data due to regulations and privacy concerns. Compared to text-based personas,  feature-based and/or event-based personas are likely easier to obtain at large scale in real applications. For example, many service providers already possess demographic features and user histories/activities, which in some way reflect the user's persona and behavioral preferences. 

Other publicly available conversation datasets, such as Reddit~\cite{mazare2018training}, Twitter~\cite{ritter2010unsupervised} or
Ubuntu\cite{lowe2015ubuntu}), do not include any real persona information other than user IDs, 
as mentioned in Section~\ref{sec:related-work:persona}. 
    
%=========================
\subsection{Lack of Behavior-Driven Dataset}
%=========================
Another challenge to exploiting personalization in conversational AI is
the lack of any dataset that leverages behavioral science that could allow one to infer and navigate specific barriers to understanding or activation that prevent the conversation from providing the assistance the user truly needs. In other words, we know that individuals have different barriers (often mental) that a conversational assistant might need to address in order to provide the expected assistance. Yet without more relevant and personal auxiliary information beyond the conversation itself or without a means for injecting a broader contextual understanding into a conversational agent, these models will always be limited in their usefulness. Exploring how these aspects of conversational AI could be addressed might be predicated upon first generating appropriate datasets on which to experiment.

The process through which the ConvAI2 dataset was generated was very
specific to the task in that the Turkers were asked to get to know each 
other. Therefore, the speakers mainly focused on sharing topics
in their assigned persona entries. 
However, there is not a real link between the persona entries
and a desired behavior that the chatbot might be intended to aid the speakers to achieve. 
The persona entries in the ConvAI2 task do not dramatically affect the way the speakers would 
interpret and respond to the conversational partner. 
For example, one Turker might share ``I work as a mathematician'',
but his/her speaking style does not necessarily reflect a precise
or critical personality. 
In addition, these responses are not intended to help the speakers 
navigate barriers to adopting certain behaviors, and therefore, would not be appropriate 
for many real-world applications that nudge people to a positive outcome, such as healthy behaviors that might be appropriate in a health-related application.

{\Raju However, such behavioral data can't be easily obtained from Turkers (e.g., ConvAI2 dataset) or social media sites (e.g., Twitter dataset), and may require guidance from highly skilled behavioral scientists. Therefore, further research is needed to address these limitations, in addition to a strong collaboration with different domain experts, as well as Turkers and volunteers to guide the process of generating rich persona-based conversational data.}

{\Raju In summary, existing datasets are not sufficient to build advanced persona-based conversational agents, in particular, in domains where more complex and multimodal data is required to engage users and drive them towards specific goals.}

%=========================
\subsection{Lack of Evaluation on Personalization}
%=========================
Current research also lacks an effective way to evaluate the quality of
personalization. Existing retrieval-based algorithms usually treat the
responses as either true or false, and the metrics used by 
generation-based algorithms typically evaluate responses based on overlapping
words in comparison with reference responses from the dataset. 
These metrics fail to address the relevance of a response to a 
speaker's and/or addressee's persona information. 
{\nc For example, when 
using conversational AI to nudge a patient to schedule an annual 
mammogram screening, one model could provide a generic
response ``Women aged 40$\sim$76 years are recommended to be screened
annually'', while another model would provide a personalized response,
such as ``Most women who are in their early 50s choose to 
screen annually to stay healthy'',
if the model is given the patient's demographic features along with her \textit{Social Proof} persona.}
Both of the example responses are correct and non-bland, but 
the second response is preferred, according to domain experts from
Lirio's Behavior Science team, as it is more tailored to the patient
to tackle her specific barrier.
Currently the only way to judge the quality of personalization of a
response is through human evaluation. It requires a significant amount
of human effort and domain expertise, which makes it very difficult to scale in real applications. 
Research efforts are needed to explore computational
methods that evaluate personalization performance at large scale. 

\section{Conclusion}
{\Raju In this paper, we: (i) reviewed the current state of the art in conversational AI focused on personalization, (ii) studied the response retrieval performance  with and without personas using two state-of-the-art methods - the Poly-Encoder and the Ranking Profile Memory Network, (iii) conducted experiments on an existing benchmark dataset, and (iv) identified limitations and provided insights into future research needs.

First, we note that the experimental results illustrate that including persona information leads to significant improvement in the performance of the conversational models. However, we have also observed the limitations of current datasets and evaluation metrics. Additionally, 
we have suggested future research directions to address several key
limitations of existing research on persona-based conversations, 
including the lack of realistic persona and behavior-driven 
conversational data, the lack of satisfactory evaluation metrics, and the difficulty of multimodal data fusion using current methods. 
\hl{In addition, we note that AI-driven algorithmic personalization and nudging
come with ethical issues. Though AI and ethics are out of the scope of 
this work, we refer interested readers to the following
\mbox{articles~\cite{Ashman-14,Hermann-22,Libai-20,Jobin-19}}}}.
\ifCLASSOPTIONcompsoc
  \section*{Acknowledgments}
\else
  \section*{Acknowledgment}
\fi

The authors would like to thank the anonymous reviewers as well as Jim Andress, Anton Dereventsov, and Clayton Webster for comments and suggestions that helped improve the quality of the paper.

\bibliographystyle{IEEEtran}
\bibliography{references}

\end{document}